%% file: main.tex
\documentclass[runningheads]{llncs}

\usepackage[english]{babel}
\usepackage[utf8]{inputenc} 
\usepackage[T1]{fontenc}    
\usepackage{hyperref}       
\usepackage{url}            
\usepackage{booktabs}       
\usepackage{amsfonts, mathrsfs, amssymb}       
\usepackage{nicefrac}       
\usepackage{microtype}      
\usepackage[export]{adjustbox}

\usepackage{graphicx}
\usepackage{subfigure,wrapfig}
\usepackage{dsfont}

\usepackage{amsmath}
\usepackage{amsthm}
\usepackage{stmaryrd}
\usepackage{xcolor}
\usepackage{breqn}
\usepackage{shortcuts}

\usepackage{algorithm}
\usepackage{algorithmic}
\usepackage[labelfont=bf]{caption}
\usepackage{etoolbox}
\usepackage{subfiles}

\begin{document}

\title{Group level MEG/EEG source imaging via optimal transport: minimum Wasserstein estimates}

\titlerunning{MWE: Minimum Wasserstein Estimates}

\author{H. Janati\inst{1}, T. Bazeille\inst{1}, B. Thirion\inst{1}, M. Cuturi\inst{2}, A. Gramfort\inst{1}}

\institute{
INRIA, CEA Neurospin, France
\and
Google and CREST ENSAE
}

%

\maketitle              
%

\input{sec/abstract}

\input{sec/introduction}

\input{sec/method}
\input{sec/experiments}

\input{sec/conclusion}

\bibliographystyle{splncs04}
\bibliography{references}

\end{document}

%% file: sec/abstract.tex
\begin{abstract}
Magnetoencephalography (MEG) and electroencephalography (EEG) are non-invasive modalities that measure the weak electromagnetic fields generated by neural activity. Inferring the location of the current sources that generated these magnetic fields is an ill-posed inverse problem known as source imaging.
When considering a group study, a baseline approach consists in carrying out the estimation of these sources independently for each subject. The ill-posedness of each problem is typically addressed using sparsity promoting regularizations. A straightforward way to define a common pattern for these sources is then to average them. A more advanced alternative relies on a joint localization of sources for all subjects taken together, by enforcing some similarity across all estimated sources. An important advantage of this approach is that it consists in a single estimation in which all measurements are pooled together, making the inverse problem better posed. Such a joint estimation poses however a few challenges, notably the selection of a valid regularizer that can quantify such spatial similarities. We propose in this work a new procedure that can do so while taking into account the geometrical structure of the cortex. We call this procedure Minimum Wasserstein Estimates (MWE). The benefits of this model are twofold. First, joint inference allows to pool together the data of different brain geometries, accumulating more spatial information. Second, MWE are defined through Optimal Transport (OT) metrics which provide a tool to model spatial proximity between cortical sources of different subjects, hence not enforcing identical source location in the group. These benefits allow MWE to be more accurate than standard MEG source localization techniques. To support these claims, we perform source localization on realistic MEG simulations based on forward operators derived from MRI scans. On a visual task dataset, we demonstrate how MWE infer neural patterns similar to functional Magnetic Resonance Imaging (fMRI) maps.

\end{abstract}
\keywords{Brain \and Inverse modeling \and EEG / MEG source imaging}

%% file: sec/introduction.tex
\section{Introduction}
\label{s:introduction}
Magnetoencephalography (MEG) measures the components of the magnetic field surrounding the head, while Electroencephalography (EEG) measures the electric potential at the surface of the scalp. Both can do so with a temporal resolution of less than a millisecond. Localizing the underlying neural activity on a high resolution grid of the cortex, a problem known as source imaging, is inherently an ``ill-posed'' linear inverse problem: Indeed, the number of potential sources is larger than the number of MEG and EEG sensors, which implies that, even in the absence of noise, different neural activity patterns could result in the same electromagnetic field measurements.

To limit the set of possible solutions, prior hypotheses on the nature of the source distributions are necessary. The minimum-norm estimates (MNE) for instance are based on $\ell_2$ Tikhonov regularization which leads to a linear solution~\cite{Hamalainen1994}. An $\ell_1$ norm penalty was also proposed by \cite{mce}, modeling the underlying neural pattern as a sparse collection of focal dipolar sources, hence their name ``Minimum Current Estimates'' (MCE). These methods have inspired a series of contributions in source localization techniques relying on noise normalization \cite{dspm,sloreta} to correct for the depth bias~\cite{depthbias} or block-sparse norms~\cite{strohmeier-etal:16,gramfort-etal:2013} to leverage the spatio-temporal dynamics of MEG signals.
While such techniques have had some success, source estimation in the presence of complex multi-dipole configurations remains a challenge. In this work we aim to leverage the anatomical and functional diversity of multi-subject datasets to improve localization results.

\paragraph{Related work.}
This idea of using multi-subject information to improve statistical estimation has been proposed before in the neuroimaging literature. In \cite{larson14} it is showed that different anatomies across subjects allow for point spread functions that agree on a main activation source but differ elsewhere. Averaging across subjects thereby increases the accuracy of source localization. On fMRI data, \cite{varoquaux11} proposed a probabilistic dictionary learning model to infer activation maps jointly across a cohort of subjects. A similar idea led \cite{gala} to introduce a Bayesian framework to account for functional intersubject variability. To our knowledge, the only contribution formulating the problem as a multi-task regression model employs a Group Lasso with an $\ell_{21}$ block sparse norm~\cite{lim17}. Yet this forces every potential neural source to be either active for all subjects or for none of them. %

\paragraph{Contribution.}
The assumption of identical functional activity across subjects is clearly not realistic. Here we investigate several multi-task regression models that relax this assumption.
One of them is the multi-task Wasserstein (MTW) model~\cite{mtw}. MTW is defined through an Unbalanced Optimal Transport (UOT) metric that promotes support proximity across regression coefficients. However, applying MTW to group level data assumes that the signal-to-noise ratio is the same for all subjects. We propose to build upon MTW and alleviate this problem by inferring estimates of both sources and noise variance for each subject. To do so, we follow similar ideas that lead to the concomitant Lasso~\cite{owen07,zhang12,ndiaye17} or the multi-task Lasso~\cite{massias18a}.


This paper is organized as follows. Section~\ref{s:sourceimaging} introduces the multi-task regression source imaging problem. Section~\ref{s:mwe} presents some background on UOT metrics  and explains how MWE are carried out. Section~\ref{s:results} presents the results of our experiments on both simulated and MEG datasets.

%% file: sec/method.tex
\label{s:methods}
\paragraph{Notation.}
We denote by $\mathds 1_p$ the vector of ones in $\bbR^p$ and by $\intset{q}$ the set $\{1, \ldots, q\}$ for any integer $q \in \bbN$. The set of vectors in $\bbR^p$ with non-negative (resp. positive) entries is denoted by $ \bbR^p_+$ (resp. $\bbR^p_{++}$).  On matrices, $\log$, $\exp$ and the division operator are applied elementwise. We use $\odot$ for the elementwise multiplication between matrices or vectors. If $\bX$ is a matrix, $\bX_{i.}$ denotes its $i^{\text{th}}$ row and $\bX_{.j}$ its $j^{\text{th}}$ column. We define the Kullback-Leibler (KL) divergence between two positive vectors by $\kl(\bx, \by) = \langle \bx , \log(\bx / \by) \rangle + \langle \by - \bx, \mathds 1_p \rangle$ with the continuous extensions  $0\log(0 / 0) = 0 $ and $0 \log(0) = 0$. We also make the convention $\bx \neq 0 \Rightarrow \kl(\bx | 0) = +\infty$. The entropy of $\bx \in \bbR^n$ is defined as $H(\bx) = - \langle \bx,\log(\bx) - \mathds 1_p \rangle $. The same definition applies for matrices with an element-wise double sum.
%
\section{Source imaging as a multi-task regression problem}
\label{s:sourceimaging}
We formulate in this section the inverse problem of interest in this paper, and recall how a multi-task formulation can be useful to carry out a joint estimation of all these parameters through regularization.
\paragraph{Source modeling.}
Using a volume segmentation of the MRI scan of each subject, the positions of potential sources are constructed as a set of coordinates uniformly distributed on the cortical surface of the gray matter. Moreover, synchronized currents in the apical dendrites of cortical pyramidal neurons are thought to be mostly responsible for MEG signals~\cite{okada93}. Therefore, the dipole orientations are usually constrained to be normal to the cortical surface. We model the current density as a set of focal current dipoles with fixed positions and orientations. The purpose of source localization is to infer their amplitudes. The ensemble of possible candidate dipoles forms the \emph{source space}.
\paragraph{Forward modeling.}
Let $n$ denote the number of sensors (EEG and/or MEG) and $p$ the number of sources.
Following Maxwell's equations, at each time instant, the measurements $\bB \in \bbR^{n}$ are a linear combination of the current density $\bx \in \bbR^p: 
\bB= \bL\bx$.
However, we observe noisy measurements $\bY \in \bbR^n$ given by:
\begin{equation}
\label{eq:inv}
 \bY = \bB + \varepsilon  = \bL \bx + \varepsilon \enspace, 
 \end{equation}
where $\varepsilon$ is the noise vector.
The linear forward operator $\bL \in \bbR^{n\times p}$ is called the \emph{leadfield} or \emph{gain matrix}, and can be computed by solving Maxwell's equations using the Boundary element method~\cite{ha87}. Up to a whitening pre-processing step, $\varepsilon$ can be assumed Gaussian distributed $\mathcal{N}(0, \sigma I_n)$.

\paragraph{Source localization.}
Source localization consists in solving in $\bx$ the inverse problem \eqref{eq:inv} which can be cast as a least squares problem:
\begin{equation}
\label{eq:leastsquares}
\bx^\star = \argmin_{\bx \in \bbR^p} \, \|\bY - \bL\bx\|^2_2 \enspace .
\end{equation}
%
Since $n \ll p$, problem \eqref{eq:leastsquares} is ill-posed and additional constraints on the solution $\bx^\star$ are necessary. When analyzing evoked responses, one can promote source configurations made of a few focal sources, e.g. using the $\ell_1$ norm. This regularization leads to problem \eqref{eq:lasso} called minimum current estimates (MCE), also known in the machine learning community as the Lasso~\cite{lasso}.
 \begin{equation}
 \label{eq:lasso}
 \bx^\star = \argmin_{\bx \in \bbR^p} \, \frac{1}{2n} \|\bY - \bL\bx\|^2_2 + \lambda\|\bx\|_1 \enspace ,
 \end{equation}
 where $\lambda > 0$ is a tuning hyperparameter.
\paragraph{Common source space.}
 Here we propose to go beyond the classical pipeline and carry out source localization jointly for $S$ subjects. First, dipole positions (features) must correspond to each other across subjects. To do so, the source space of each subject is mapped to a high resolution average brain using morphing where the sulci and gyri patterns are matched in an auxiliary spherical inflating of each brain surface \cite{morphing}. The resulting leadfields $\bL^{(1)}, \dots, \bL^{(S)}$ have therefore the same shape $(n\times p)$ with aligned columns.
 \paragraph{Multi-task framework.}
 Jointly estimating the current density $\bx^{(s)}$ of each subject $s$ can be expressed as a multi-task regression problem where some coupling prior is assumed on $\bx^{(1)}, \dots, \bx^{(S)}$ through a penalty $\Omega$:
 \begin{equation}
 \label{eq:multitask}
    \min_{\bx^{(1)}, \dots, \bx^{(S)} \in \bbR^p} \,
    \frac{1}{2n} \sum_{s=1}^S \|\bY^{(s)} - \bL^{(s)}\bx^{(s)}\|^2_2 \,
    + \, \Omega(\bx^{(1)}, \dots, \bx^{(S)}) \enspace.
 \end{equation}
Following the work of \cite{mtw}, we propose to define $\Omega$ using an UOT metric.

%
\section{Minimum Wassertein Estimates}
\label{s:mwe}
We start this section with background material on UOT. Consider the finite metric space $(E, d)$  where each element of $E = \{1, \dots, p\}$ corresponds to a vertex of the source space. Let $\bM$ be the matrix where $\bM_{ij}$ corresponds to the geodesic distance between vertices $i$ and $j$. Kantorovich \cite{doklady} defined a distance for normalized histograms (probability measures) on $E$. However, it can easily be extended to non-normalized measures by relaxing marginal constraints~\cite{chizat:17}. 
\paragraph{Marginal relaxation.}
Let $\ba, \bb$ be two normalized histograms on $E$. Assuming that transporting a fraction of mass $\bP_{ij}$ from $i$ to $j$ is given by $\bP_{ij} \bM_{ij}$, the total cost of transport is given by $\langle \bP, \bM\rangle = \sum_{ij} \bP_{ij} \bM_{ij}$. Minimizing this total cost with respect to $\bP$ must be carried out on the set of feasible transport plans with marginals $\ba$ and $ \bb$. The (normalized) Wasserstein-Kantorovich distance reads:
\begin{equation}
\label{eq:wasserstein}
\text{WK}(\ba, \bb) = \min_{\substack{\bP \in {\bbR_+}^{p\times p} \\ \bP\mathds 1 = \ba, \bP^\top \mathds 1 = \bb}} \, \langle \bP, \bM \rangle \enspace.
\end{equation}
In practice, if $\ba$ and $ \bb$ are positive and normalized current densities,  WK$(\ba, \bb)$  will quantify the geodesic distance between their supports along the curved geometry of the cortex. This property makes OT metrics adequate for assessing the proximity of functional patterns across subjects.
To allow $\ba, \bb$ to be non-normalized, the marginal constraints in \eqref{eq:wasserstein} can be relaxed using a KL divergence:
 \begin{equation}
\label{eq:relaxed-marginals}
\min_{\bP \in {\bbR_+}^{p\times p}} \, \langle \bP, \bM \rangle + \gamma \kl(\bP\mathds 1 | \ba) + \gamma \kl(\bP^\top \mathds 1 | \bb) \enspace,
\end{equation}
where $\gamma > 0$ is a hyperparameter that enforces a fit to the marginals.
\paragraph{Entropy regularization.}
Entropy regularization was introduced by \cite{cuturi:13} to propose a faster and more robust alternative to the direct resolution of the linear programming problem \eqref{eq:wasserstein}. Formally, this amounts to minimizing the loss $ \langle \bP, \bM \rangle - \varepsilon H(\bP) $ where $\varepsilon > 0$ is a tuning hyperparameter. This penalized loss function can be written: $\varepsilon \kl(\bP, e^{- \frac{\bM}{\varepsilon}})$ up to a constant \cite{benamou:15}. Combining entropy regularization with marginal relaxation in \eqref{eq:relaxed-marginals}, we get the unbalanced Wasserstein distance as introduced by \cite{chizat:17}:
 \begin{equation}
\label{eq:unbalanced-wasserstein}
W(\ba, \bb) = \min_{\bP \in {\bbR_+}^{p\times p}} \,\varepsilon \kl(\bP| e^{- \frac{\bM}{\varepsilon}}) + \gamma \kl(\bP\mathds 1 | \ba) + \gamma \kl(\bP^\top \mathds 1 | \bb) \enspace,
\end{equation}
\paragraph{Generalized Sinkhorn.}
Problem \eqref{eq:unbalanced-wasserstein} can be solved as follows.
Let $\bK = e^{- \frac{\bM}{\varepsilon}}$ and $\psi = \gamma / (\gamma + \epsilon)$. Starting from two vectors $\bu, \bv$ set to $\mathds 1$ and iterating the scaling operations $\bu \leftarrow \left(\ba / \bK\bv\right)^\psi$ , $\bv \leftarrow \left(\bb / \bK^\top \bu\right)^\psi$ until convergence, the minimizer of \eqref{eq:unbalanced-wasserstein} can be computed as $ P^{\star} = (\bu_i\bK_{ij}\bv_j)_{i, j \in \intset{p}}$. This algorithm is a generalization of the Sinkhorn algorithm \cite{knopp}. Since it involves matrix-matrix operations, it benefits from parallel hardware, such as GPUs. 
\paragraph{Extension to $\bbR^p$.}
We extend next the Wasserstein distance to signed measures. We adopt a similar idea to what was suggested in~\cite{mainini,sturm,mtw} using a decomposition into positive and negative parts, $\bx^{(s)} = \bx^{(s)_+} - \bx^{(s)_-}$ where $\bx^{(s)_+} = \max(\bx^{(s)_+}, 0)$ and $\bx^{(s)_-} = \max(- \bx^{(s)_+}, 0)$.  For any vectors $\ba, \bb \in \bbR^p$, we define the generalized Wasserstein distance as:
 \begin{equation}
\label{eq:signed-wasserstein}
\widetilde{W}(\ba, \bb) \eqdef W(\ba^+, \bb^+) + W(\ba^-, \bb^-) \enspace.
\end{equation}
Note that $W(\boldsymbol{0}, \boldsymbol{0}) = 0$ (see \cite{mtw} for a proof), thus on positive measures $\widetilde{W} = W$.
For the sake of convenience, we refer to $\widetilde{W}$ in \eqref{eq:signed-wasserstein} by the Wasserstein distance, even though it does not verify indiscernability. In practice, this extension allows to compare current dipoles across subjects according to their polarity which could be either towards the deep or superficial layers of the cortex.
%
 \paragraph{The MTW model.}
 The multi-task Wasserstein model is the specific case of  \eqref{eq:multitask} with a penalty $\Omega$ promoting both sparsity and supports' proximity:
   \begin{equation}
 \label{eq:mtw}
\Omega_{\text{MTW}}(\bx^{(1)}, \dots, \bx^{(S)}) \eqdef \mu \min_{\bxbar  \in \bbR^p}  \frac{1}{S}\sum_{s=1}^S \widetilde{W}(\bx^{(s)}, \bxbar) + \lambda \|\bx^{(s)}\|_1 \enspace,
 \end{equation}
 where $\mu, \lambda \geq 0$ are tuning hyperparameters.
 The OT term in \eqref{eq:mtw} can be seen as a spatial variance. Indeed, the minimizer $\bxbar$ corresponds to the Wasserstein barycenter with respect to the distance $\widetilde{W}$.
  \paragraph{Minimum Wasserstein Estimates.}
  One of the drawbacks of MTW is that $\lambda$ is common to all subjects. Indeed, the loss considered in MTW implicitly assumes that the level of noise is the same across subjects. Following the work of \cite{ndiaye17} on the smoothed concomitant Lasso, we propose to extend MTW by inferring the specific noise standard deviation  ${\sigma^{(s)}}$ along with the regression coefficient $\bx^{(s)}$ of each subject. This allows to scale the weight of the $\ell_1$ according to the level of noise. The Minimum Wasserstein Estimates (MWE) model reads:
  \begin{equation}
\label{eq:mwe}
\min_{\substack{ \bx^{(1)}, \dots, \bx^{(S)} \in \bbR^p \\ 
	\sigma^{(1)}, \dots, \sigma^{(S)} \in [\sigma_0, +\infty] }} \,  \sum_{s=1}^S \frac{1}{2n  \sigma^{(s)}}\|\bY^{(s)} - \bL^{(s)}\bx^{(s)}\|^2_2 \, +\frac{ \sigma^{(s)}}{2} + \, \Omega_{\text{MTW}}(\bx^{(1)}, \dots, \bx^{(S)}) \enspace,
\end{equation}
where $\sigma_0$ is a pre-defined constant. This lower bound  constraint avoids numerical issues when $\lambda \to 0$ and therefore the standard deviation estimate also tends to 0. In practice $\sigma_0$ can be set for example using prior knowledge on the variance of the data or as a small fraction of the initial estimate of the standard deviation $\sigma_0 = \alpha \min_s \frac{\| \bY^{(s)}\|}{\sqrt{n}}$. In practice we adopt the second option and set $\alpha = 0.01$. 
\begin{algorithm}[t]
	\caption{MWE algorithm}
	\label{alg:alt}
	\begin{algorithmic}
		\STATE {\bfseries Input:}  $\sigma_0$, $\mu, \epsilon , \gamma, \lambda$ and cost matrix $\bM$. data $(\bY^{(s)})_s (\bL^{(s)})_s$.
		\STATE {\bfseries Output:} MWE: $(\bx^{(s)})$, minimizers of \eqref{eq:mwe}.
		
		\REPEAT
		\FOR{$s=1$ {\bfseries to} $S$}
		\STATE Update $\bx^{(s)+}$ with proximal coordinate descent to solve \eqref{eq:cd}.
		\STATE Update $\bx^{(s)-}$ with proximal coordinate descent to solve \eqref{eq:cd}.
		\STATE Update $\sigma^{(s)}$ with \eqref{eq:sigma}.
		\ENDFOR
		\STATE Update  left marginals $\bm^{(1)+}, \dots, \bm^{(S)+}$  and $ \bxbar^+$ with generalized Sinkhorn.
		\STATE Update left marginals $\bm^{(1)-} \dots, \bm^{(S)-}$  and $ \bxbar^-$ with generalized Sinkhorn.
		\UNTIL{convergence}
	\end{algorithmic}
\end{algorithm}

 \paragraph{Algorithm.} 
 By combining \eqref{eq:unbalanced-wasserstein}, \eqref{eq:signed-wasserstein}
and \eqref{eq:mwe}, we obtain an objective function taking as arguments $\left((\bx^{(s)+})_s, (\bx^{(s)-})_s, (\bP^{(s)+})_s, (\bP^{(s)-})_s, \bxbar^+, \bxbar^-, (\sigma^{(s)})_s \right)$. This function restricted to all parameters except $(\sigma^{(s)})_s$ is jointly convex \cite{mtw}. Moreover, each $\sigma^{(s)}$ is only coupled with the variable $\bx^{(s)}$. The restriction on every pair $(\bx^{(s)}, \sigma^{(s)})$ is also jointly convex \cite{ndiaye17}. Thus the problem is jointly convex in all its variables. We minimize it by alternating optimization. To justify the convergence of such an algorithm, one needs to notice that the non-smooth $\ell_1$ norms in the objective are separable~\cite{Tseng01}.
The update with respect to each $\sigma^{(s)}$ is given by solving the first order optimality condition (Fermat's rule): 
  \begin{equation}
\label{eq:sigma}
\sigma^{(s)} \leftarrow  \frac{ \|\bY^{(s)} - \bL^{(s)}\bx^{(s)}\|_2}{\sqrt{n}} \wedge \sigma_0 \enspace,
\end{equation}
 which also corresponds to the empirical estimator of the standard deviation when the constraint is not active. To update the remaining variables, we follow the same optimization procedure laid out in \cite{mtw} and adapted to MWE in Algorithm~\ref{alg:alt}. Briefly, let $\bm^{(s)+} \eqdef \bP^{(s)+}\mathds 1$ (resp. $\bm^{(s)+} \eqdef \bP^{(s)+}\mathds 1 $), when minimizing with respect to one $\bx^{(s)+}$ (resp. $\bx^{(s)-}$), the resulting problem can be written (dropping the exponents for simplicity):
  \begin{equation}
\label{eq:cd}
\min_{\bx \in \bbR^p_+}  \frac{1}{2n} \|\bY - \bL\bx\|_2^2 +  \frac{\mu \gamma}{S} ( \langle \bx, \mathds 1 \rangle - \langle \log(\bx), \bm \rangle ) + \lambda \sigma \|\bx \|_1 \enspace,
\end{equation}
which can be solved using proximal coordinate descent \cite{fercoq}. 
Note that the additional inference of a specific $\sigma^{(s)}$ for each subject allows to scale the Lasso penalty depending on their particular level of noise. 
The final update with respect to $((\bP^{(s)+})_s, (\bP^{(s)-})_s, \bxbar^+, \bxbar^-)$ can be cast as two Wasserstein barycenter problems, carried out using  generalized Sinkhorn iterations \cite{chizat:17}. Note that we do not need to compute the transport plans $P^{(s)}$ since inferring every source estimate $\bx$ only requires the knowledge of the left marginal $\bm = \bP\mathds 1$ which does not require storing $\bP$ in memory.

%% file: sec/experiments.tex
\section{Experiments}
\label{s:results}
\paragraph{Benchmarks: Dirty models and Multi-level Lasso.}
As discussed in introduction, standard sparse source localization solvers are based on an $\ell_1$ norm regularization, applied to the data of each subject independently. We use the independent Lasso estimator as a baseline. We compare MWE to the Group-Lasso estimator \cite{grouplasso,argyriou-etal:06} which was proposed in this context to promote functional consistency across subjects~\cite{lim17}. It falls in the multi-task framework of \eqref{eq:multitask} where the joint penalty is defined through an $\ell_{21}$ mixed norm  $ \|\bX\|_{21} = \sum_{j=1}^{p} \sqrt{\sum_{s=1}^S {\bx^{(s)}_j}^2} $ where $\bX~=~(\bx^{(s)}_j )_{(j, s)}~\in~\bbR^{p \times S}$. We also evaluate the performance of more flexible block sparse models where only a fraction of the source estimates are shared across all tasks: Dirty models \cite{dirty} and the multivel lasso \cite{multilevel}. In Dirty models source estimates are written as a sum of two parts which are penalized with different norms. One common to all subjects (penalty $\ell_{21}$) and one specific for each subject (penalty $\ell_1$). The Multi-level Lasso (MLL) \cite{multilevel} applies the same idea using instead a product decomposition and a Lasso penalty on both parts. We also compare MWE with MTW to evaluate the benefits of inferring noise levels adaptively.

\begin{figure}[!t]
	\begin{minipage}{0.3\linewidth}
		\includegraphics[width=\linewidth]{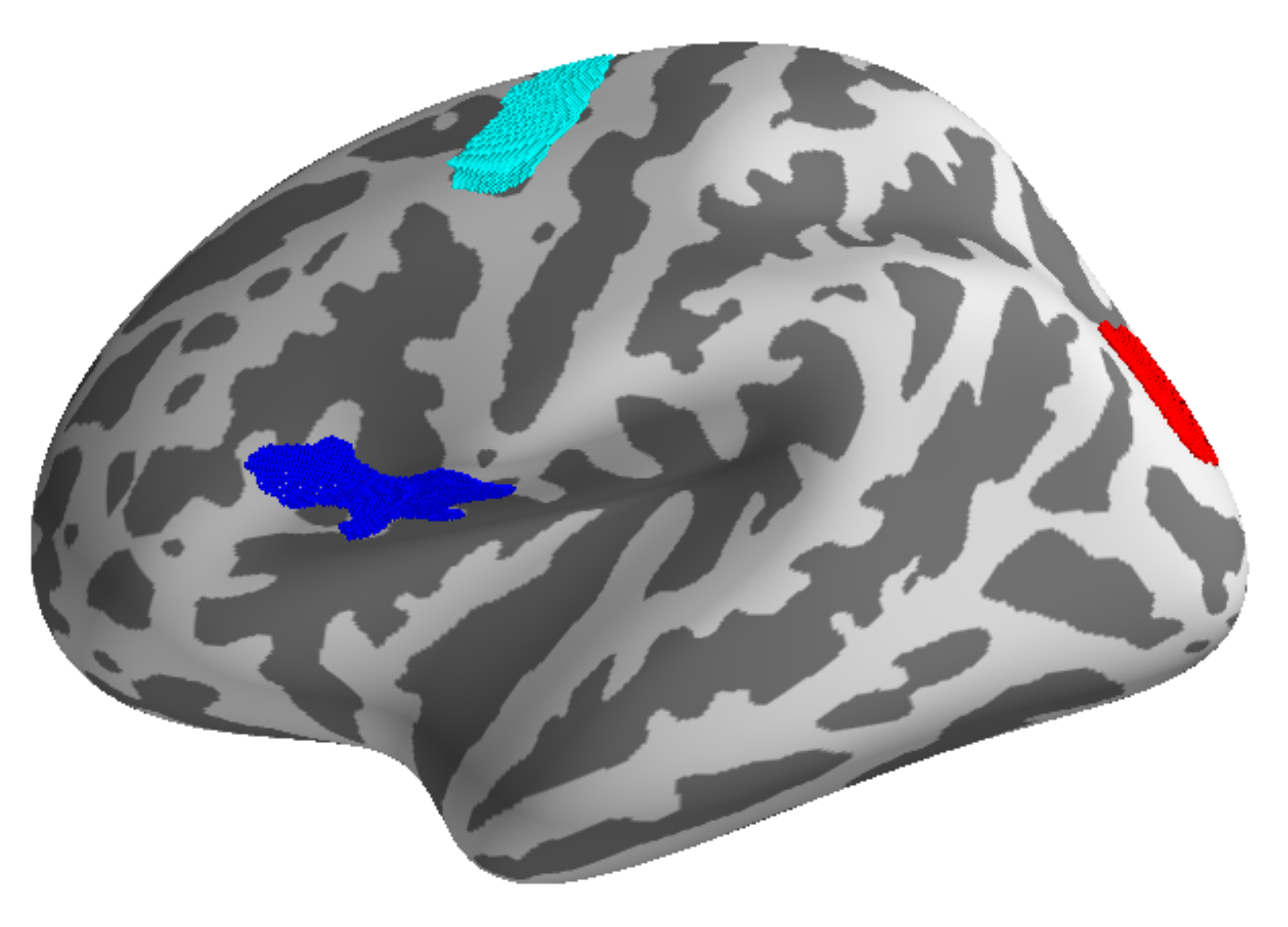}
	\end{minipage}
	\begin{minipage}{0.3\linewidth}
		\includegraphics[width=\linewidth]{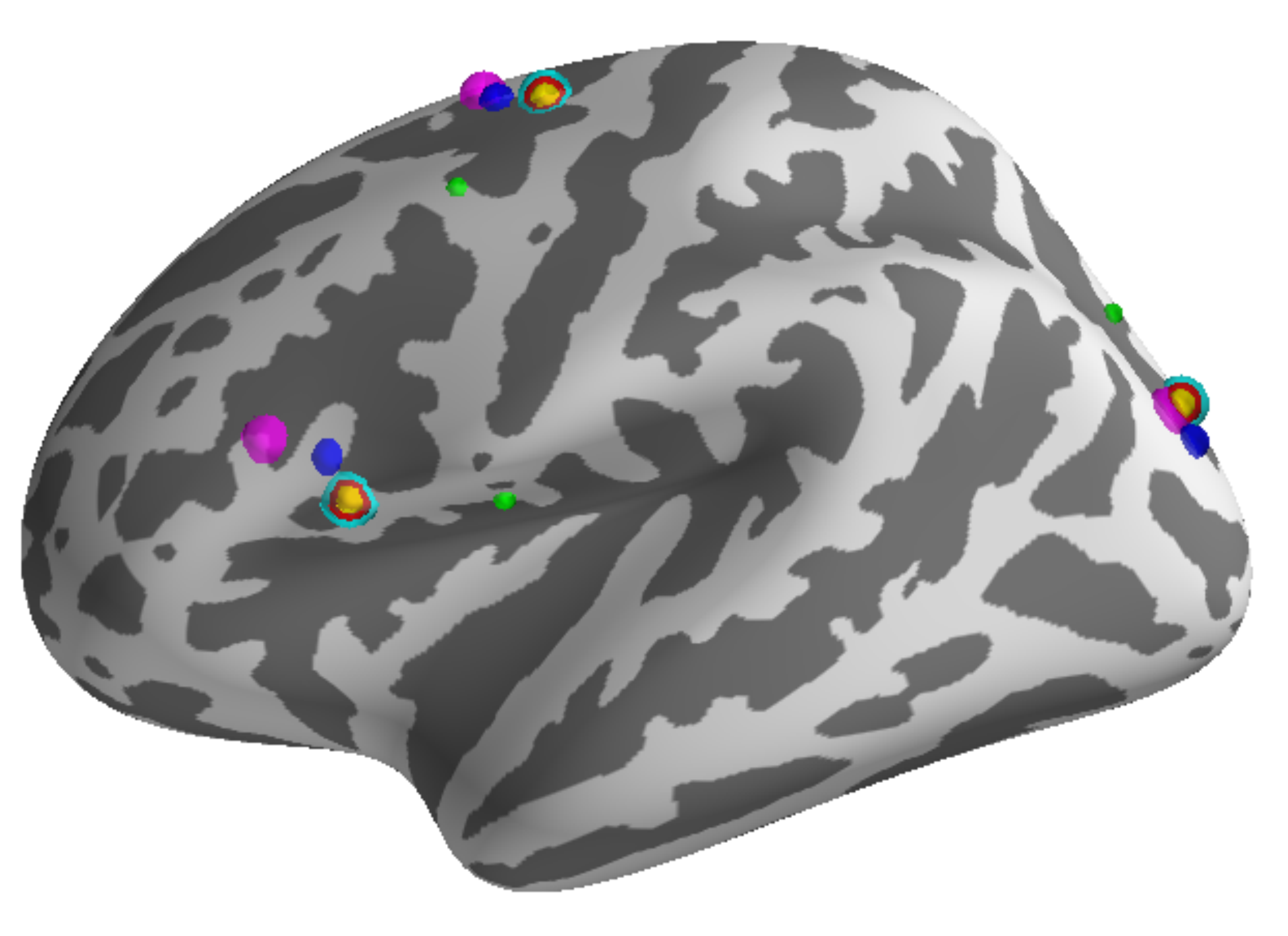}
	\end{minipage}
	\begin{minipage}{0.37\linewidth}
		\caption{\textbf{Left:} 3 labels from the aparc.a2009s parcellation. \textbf{Right:}  Simulated activations for $S=6$ subjects. Each color corresponds to a subject. Different radiuses are used to distinguish overlapping sources.\label{f:labels}}
	\end{minipage}
\end{figure}
\paragraph{Simulation data and MEG/fMRI datasets.}
We use the public dataset DS117~\cite{ds117} which provides MEG, EEG and fMRI data of 16 healthy subjects to whom were presented images of famous, unfamiliar and scrambled faces. Using the MRI scan of each subject, we compute a source space and its associated leadfield comprising around 2500 sources per hemisphere~\cite{mne}. Keeping only MEG gradiometer channels, we have $n = 204$ observations per subject.

For realistic data simulation, we use the actual leadfields from all subjects, yet restricted to the left hemisphere. We thus have 16 leadfields with $p=2500$. We simulate an inverse solution $\bx^{s}$ with $q$ sources ($q$-sparse vector) by randomly selecting one source per label among $q$ pre-defined labels using the \emph{aparc.a2009s} parcellation of the Destrieux atlas. To model functional consistency, 50\% of the subjects share sources at the same locations, the remaining 50\% have sources randomly generated in the same labels (see Figure~\ref{f:labels}). Their amplitudes are taken uniformly between 20 and 30~nAm. Their sign is taken at random with a Bernoulli distribution (0.5) for each label. We simulate $\bY$ using the forward model with a variance matrix $\sigma I_n$. We set $\sigma$ so as to have an average signal-to-noise ratio across subjects equal to 4 (SNR$\eqdef \sum_{s=1}^S \frac{\|\bL^{(s)}\bx^{(s)}\|}{ S\sigma}$).
\begin{figure}[!t]
	\includegraphics[width=\linewidth]{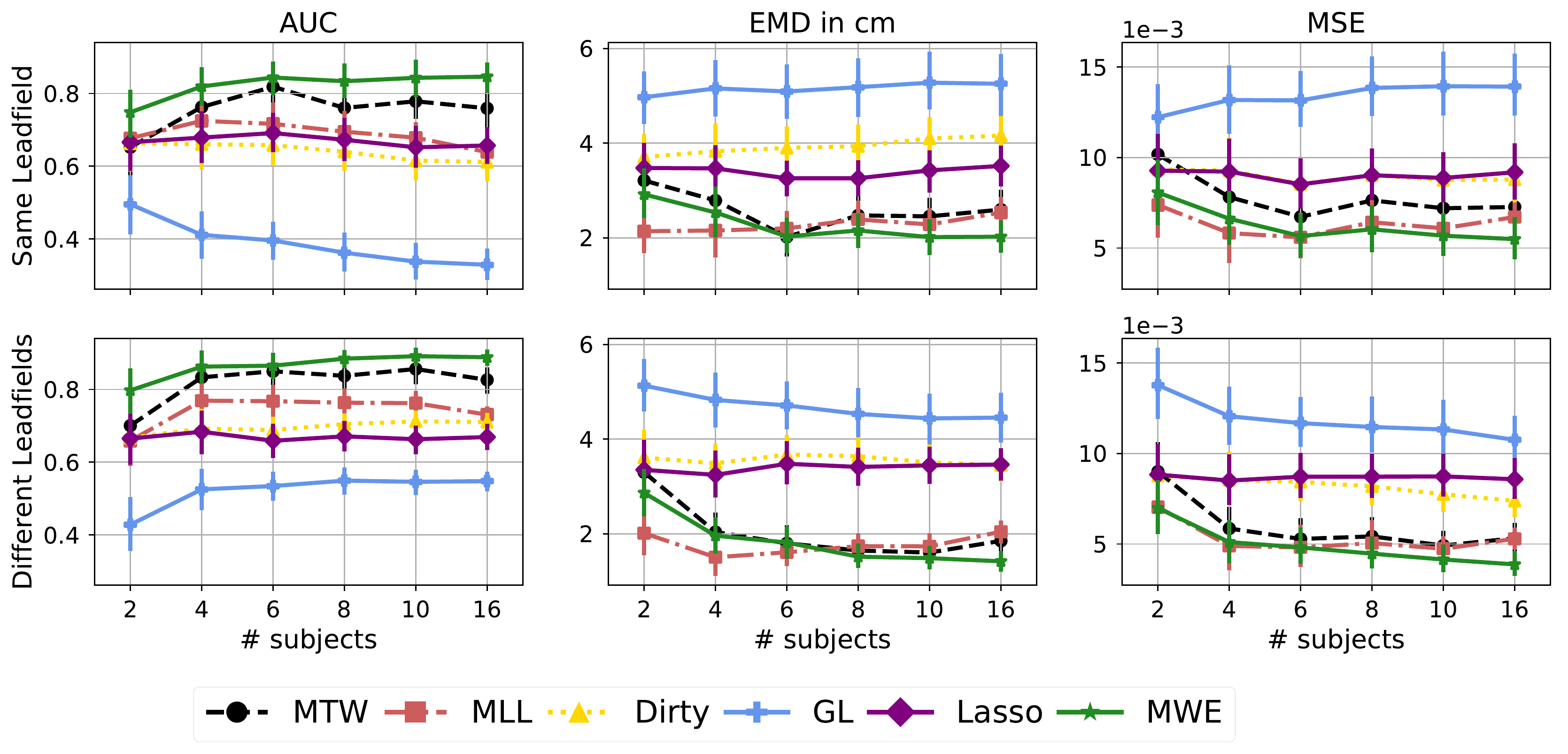}
	\caption{Performance of different models over 30 trials in terms of AUC, EMD and MSE using the same leadfield for all subjects (randomly selected in each trial) (\textbf{top}) and specific leadfields (\textbf{bottom}). \label{f:anatomies}}
\end{figure}
%
We evaluate the performance of all models knowing the ground truth by comparing the best estimates in terms of three metrics: the mean squared error (MSE) to quantify accuracy in amplitude estimation, AUC and a generalized Earth mover distance (EMD) to assess supports estimation. We generalize the PR-AUC (Area under the curve Precision-recall) by defining AUC$(\hat{\bx}, \bx^{\star}) = \frac{1}{2} 
\text{PR-AUC}(\hat{\bx}^+, \bx^{\star +}) +\frac{1}{2}  \text{PR-AUC}(\hat{\bx}^-, \bx^{\star -})$ where PR-AUC is computed between the estimated coefficients and the true supports. We compute EMD between normalized values of sources: EMD$(\hat{\bx}, \bx^{\star}) =  \frac{1}{2} 
\text{WK} (\frac{\hat{\bx}^+}{\hat{\bx}^+\mathds 1}, \frac{\bx^{\star +}}{ \bx^{\star +}\mathds 1} ) + \frac{1}{2}  \text{WK} (\frac{\hat{\bx}^-}{\hat{\bx}^-\mathds 1}, \frac{\bx^{\star -}}{ \bx^{\star -}\mathds 1} )$. Since $\bM$ is expressed in centimeters, WK can be seen as an expectation of the geodesic distance between sources. 
The mean across subjects is reported for all metrics. 
\paragraph{Simulation results.}
We set the number of sources to 3 and vary the number of subjects under two conditions: (1) using one leadfield for all subjects, (2) using individual leadfields. Each model is fitted on a grid of hyperparameters and the best AUC/MSE/EMD scores are reported. We perform 30 different trials (with different true activations and noise, different common leadfield for condition (1)) and report the mean within a 95\% confidence interval in Figure \ref{f:anatomies}.

Various observations can be made. The Group Lasso performs poorly -- even compared to independent Lasso -- which is expected since sources are not common for all subjects. Non-convexity allows MLL to be very effective with less than 2-4 subjects. Its performance yet degrades with more subjects. OT-based models (MWE and MTW) however benefit from the presence of more subjects by leveraging spatial proximity. They reduce the average error distance from 4 cm (Lasso) to less than 1 cm and reach an AUC of 0.9. One can also observe that the estimation of the noise standard deviation in the MTW model does improve performance. Finally, we can appreciate the improvement of multi-task models when increasing the number subjects, especially when using different leadfield matrices. We argue that the different folding patterns of the cortex across subjects lead to different dipole orientations thereby increasing the chances of source identification.
\paragraph{Results on MEG/fMRI data}
%
\begin{figure}[!t]
	\begin{minipage}{0.24\linewidth}
		\includegraphics[trim={0.5cm 0.5cm 0.5cm 4cm}, clip, width=\linewidth]{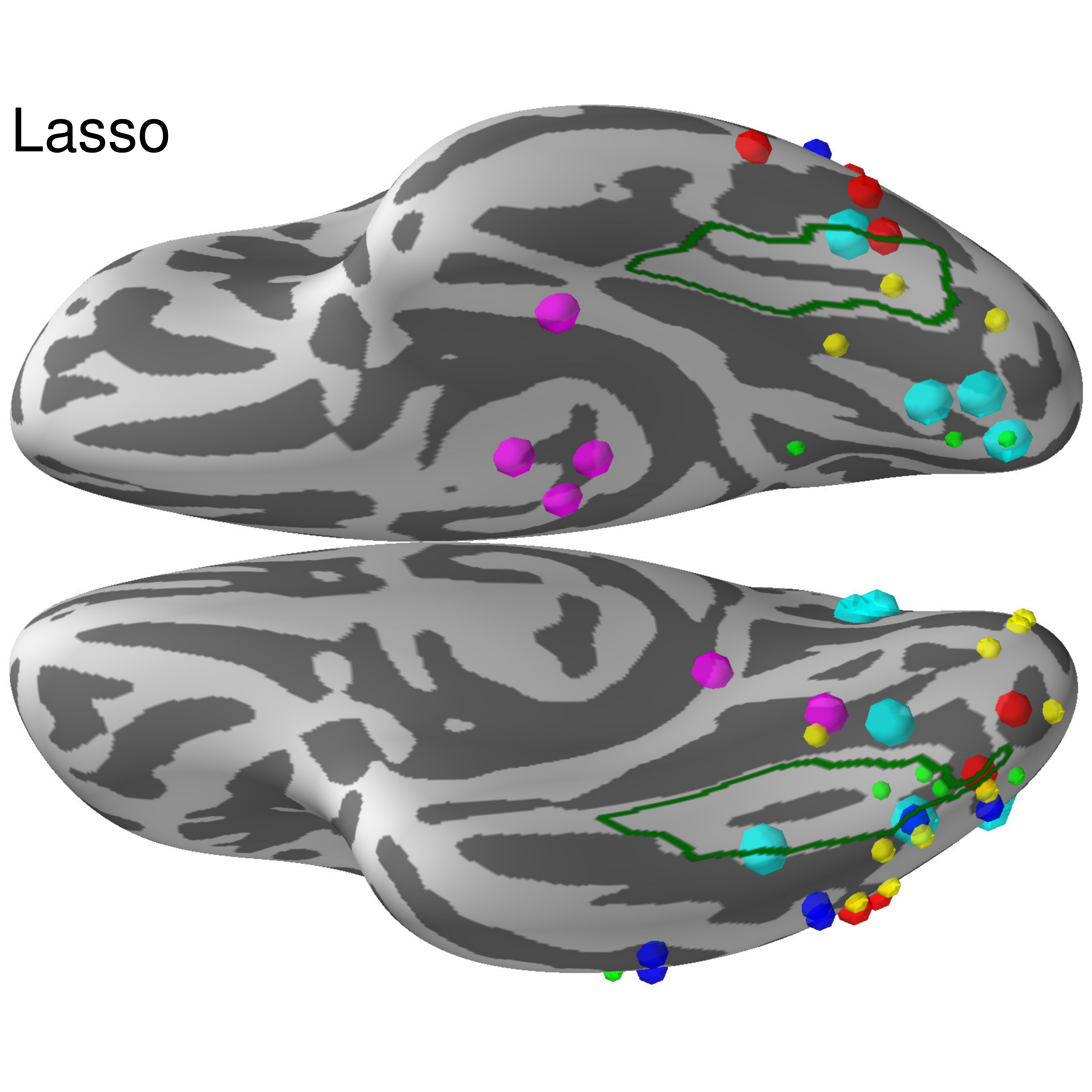}
	\end{minipage}
	\begin{minipage}{0.24\linewidth}
		\includegraphics[trim={0.5cm 0.5cm 0.5cm 4cm}, clip, width=\linewidth]{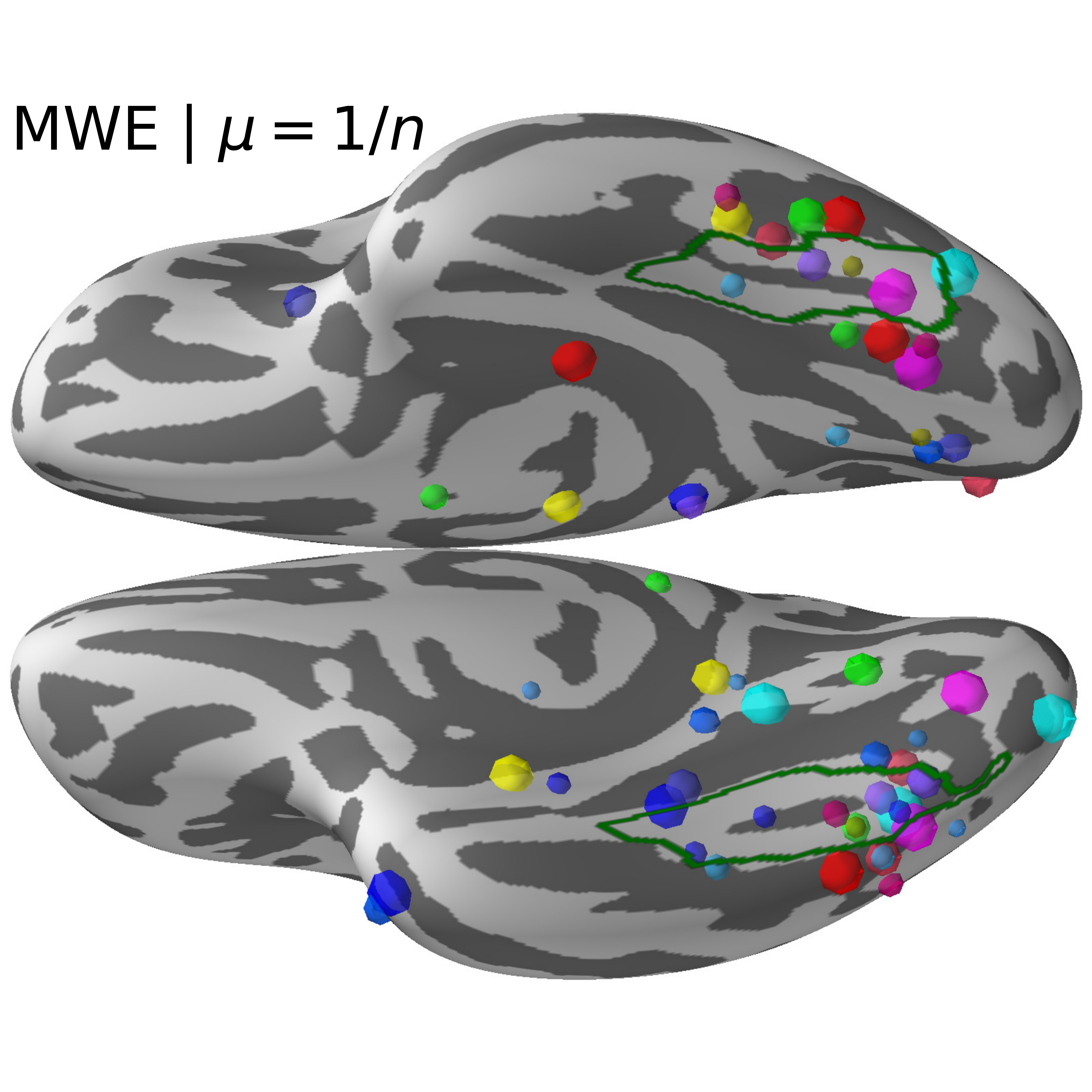}
	\end{minipage}
	\begin{minipage}{0.25\linewidth}
		\includegraphics[trim={0.5cm 0.5cm 0.5cm 4cm}, clip, width=\linewidth]{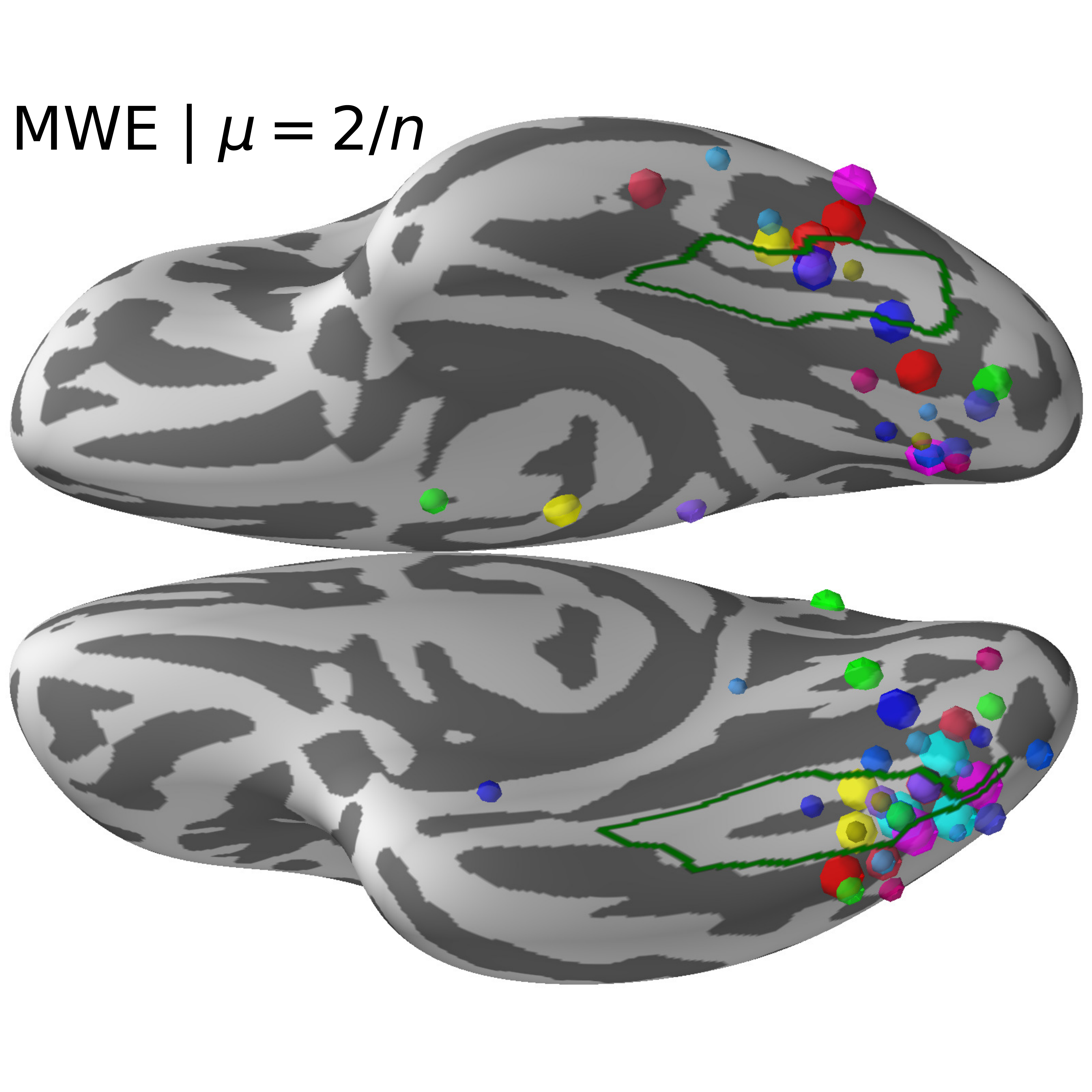}
	\end{minipage}
	\begin{minipage}{0.25\linewidth}
		\includegraphics[trim={0.5cm 0.5cm 0.5cm 4cm}, clip, width=\linewidth]{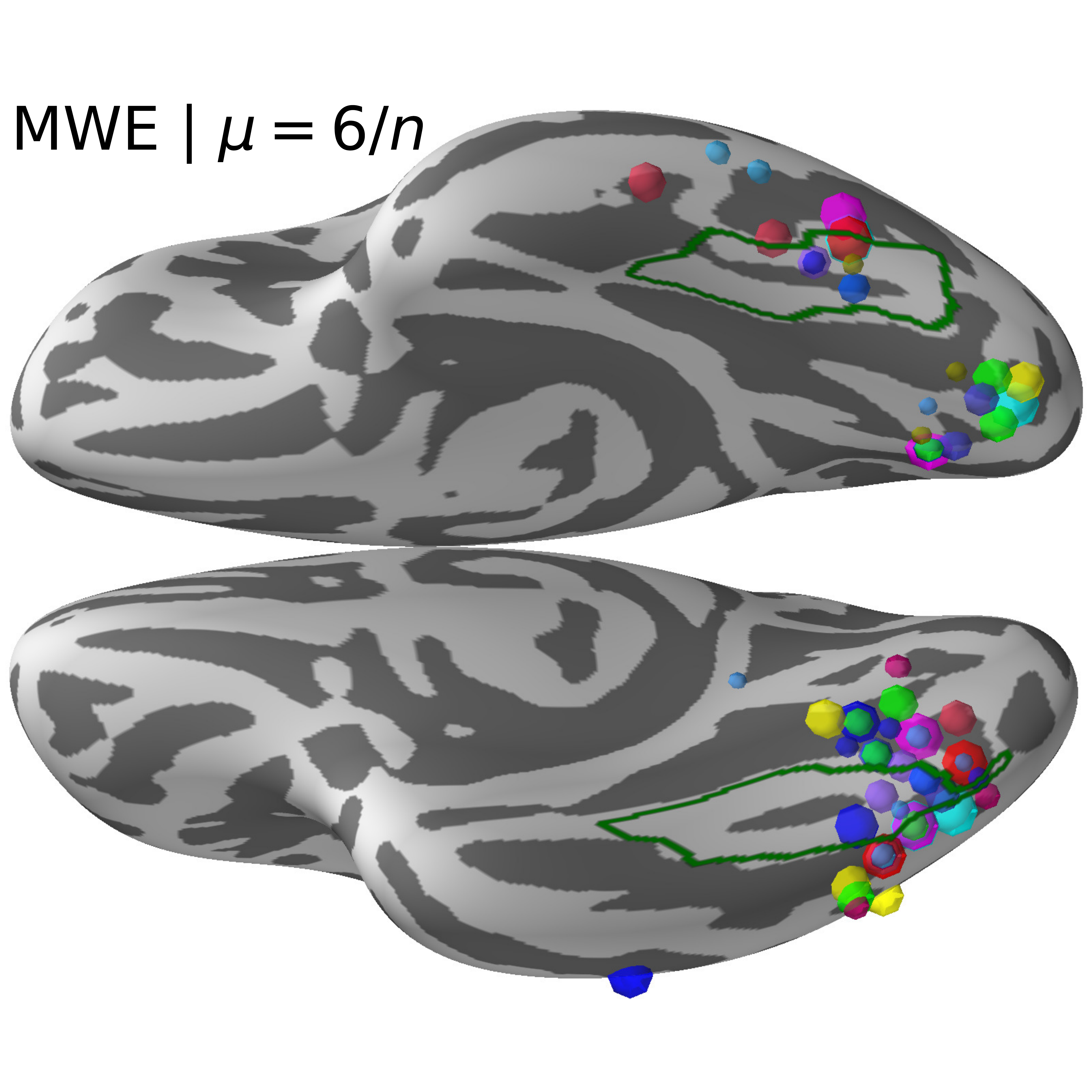}
	\end{minipage}
	\caption{Support of source estimates. Each color corresponds to a subject. Different radiuses are displayed for a better distinction of sources.  The  fusiform gyrus is highlighted in green. Increasing $\mu$ promotes functional consistency across subjects.
		\label{f:allsubjects}}
\end{figure}
The fusiform face area specializes in facial recognition and activates around 170ms after stimulus~\cite{kanwisher1997,wakeman11}. To study this response, we perform MEG source localization using Lasso and MWE. We pick the time point with the peak response for each subject within the interval 150-200 ms after visual presentation of famous faces. For both models, we select the smallest $\ell_1$ tuning parameter $\lambda$ for which less than 10 sources are active for each subject. Figure~\ref{f:allsubjects} shows how UOT regularization favors activation in the ventral pathway of the visual cortex. 
\begin{figure}[!t]
	\includegraphics[width=\linewidth]{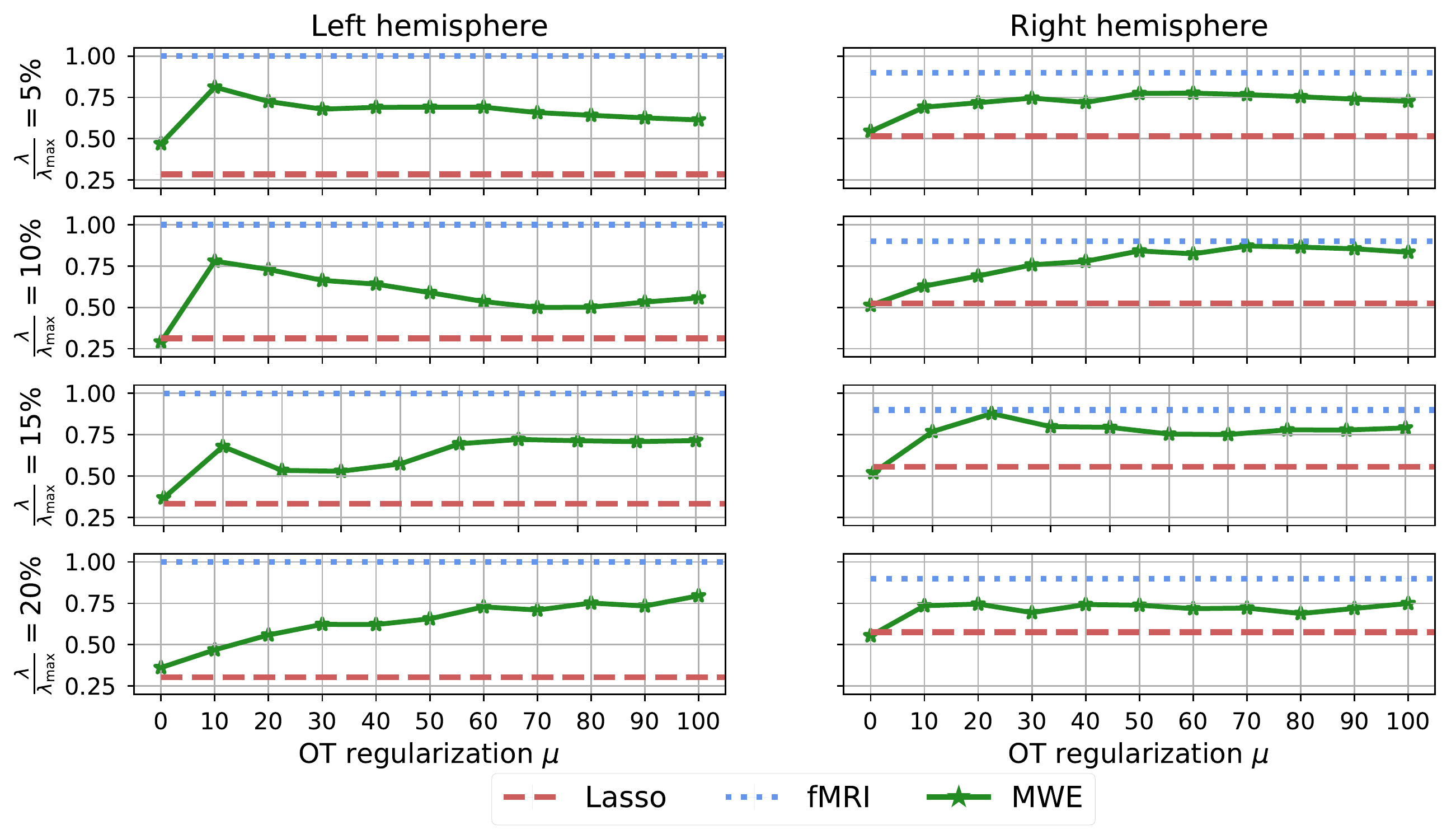}
	\caption{Ratio of maximum absolute amplitude in the fusiform gyrus over  maximum absolute amplitude in the hemisphere. The mean across the 16 subjects is reported for different $\ell_1$ norm regularization weights $\lambda$. 
		\label{f:ffg}}
\end{figure}
\begin{figure}[!tb]
	\begin{minipage}{0.32\linewidth}
		\includegraphics[trim={0.5cm 0.5cm 0.5cm 4cm}, clip, width=\linewidth]{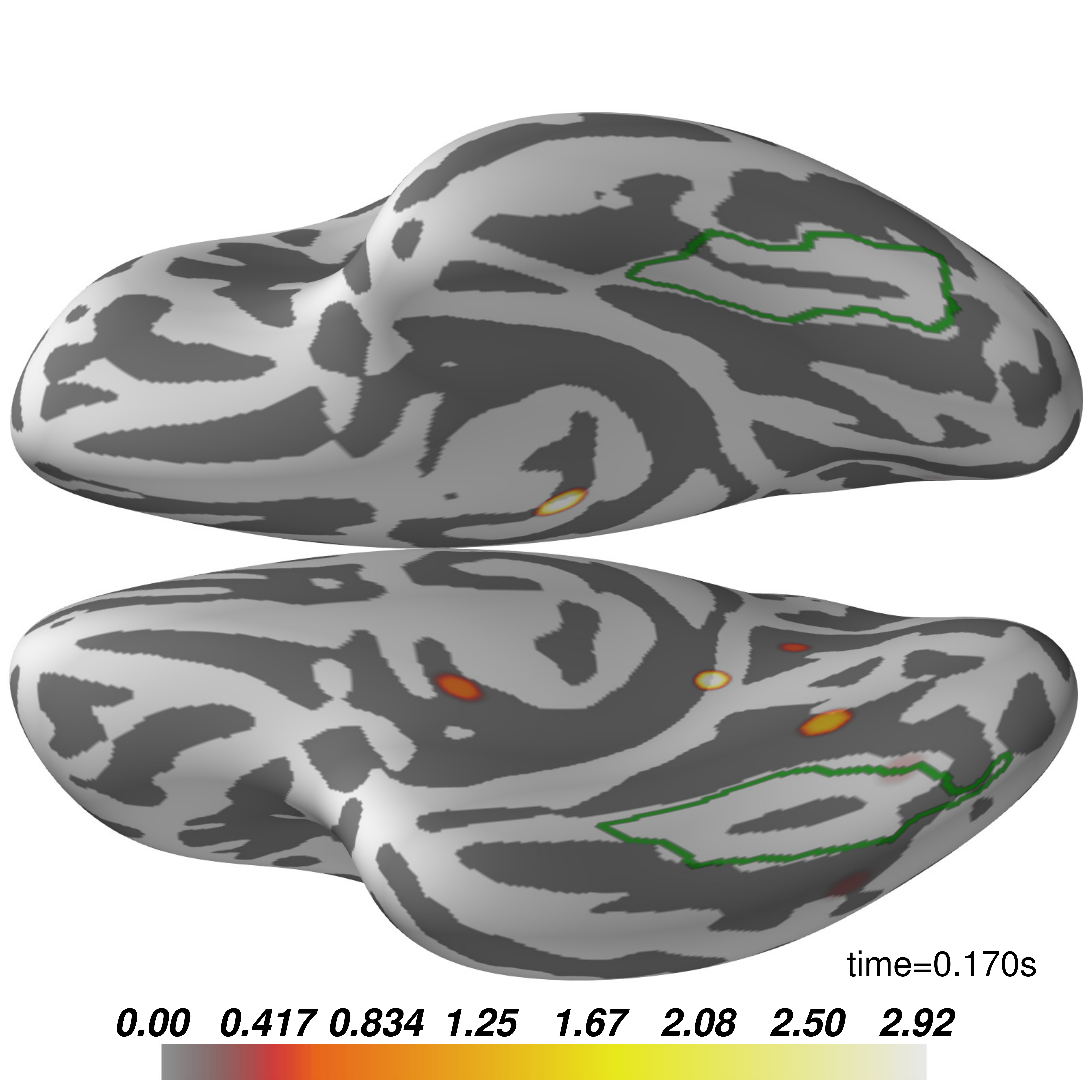}
	\end{minipage}
	\begin{minipage}{0.32\linewidth}
		\includegraphics[trim={0.5cm 0.5cm 0.5cm 4cm}, clip, width=\linewidth]{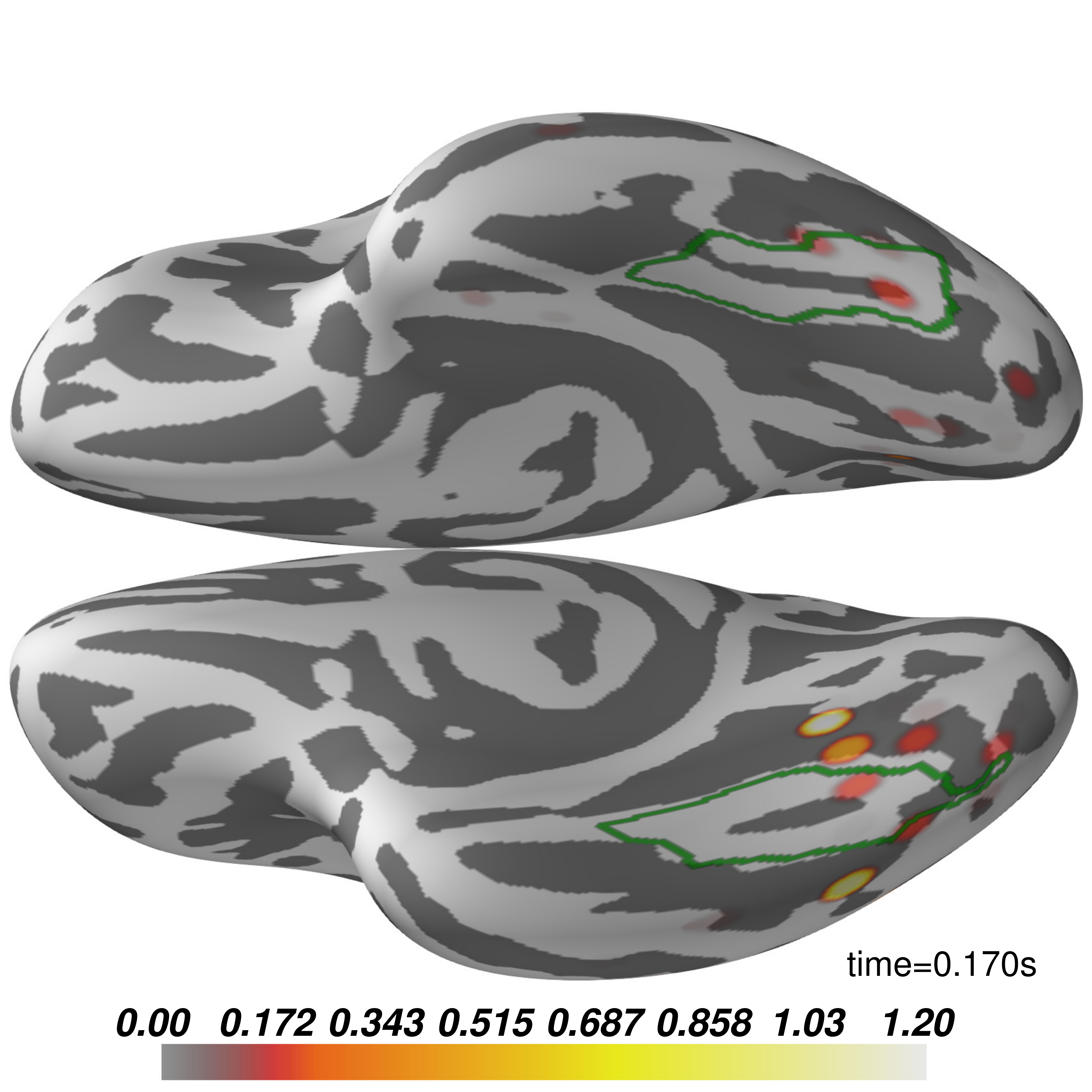}
	\end{minipage}
	\begin{minipage}{0.32\linewidth}
		\includegraphics[trim={0.5cm 0.5cm 0.5cm 4cm}, clip, width=\linewidth]{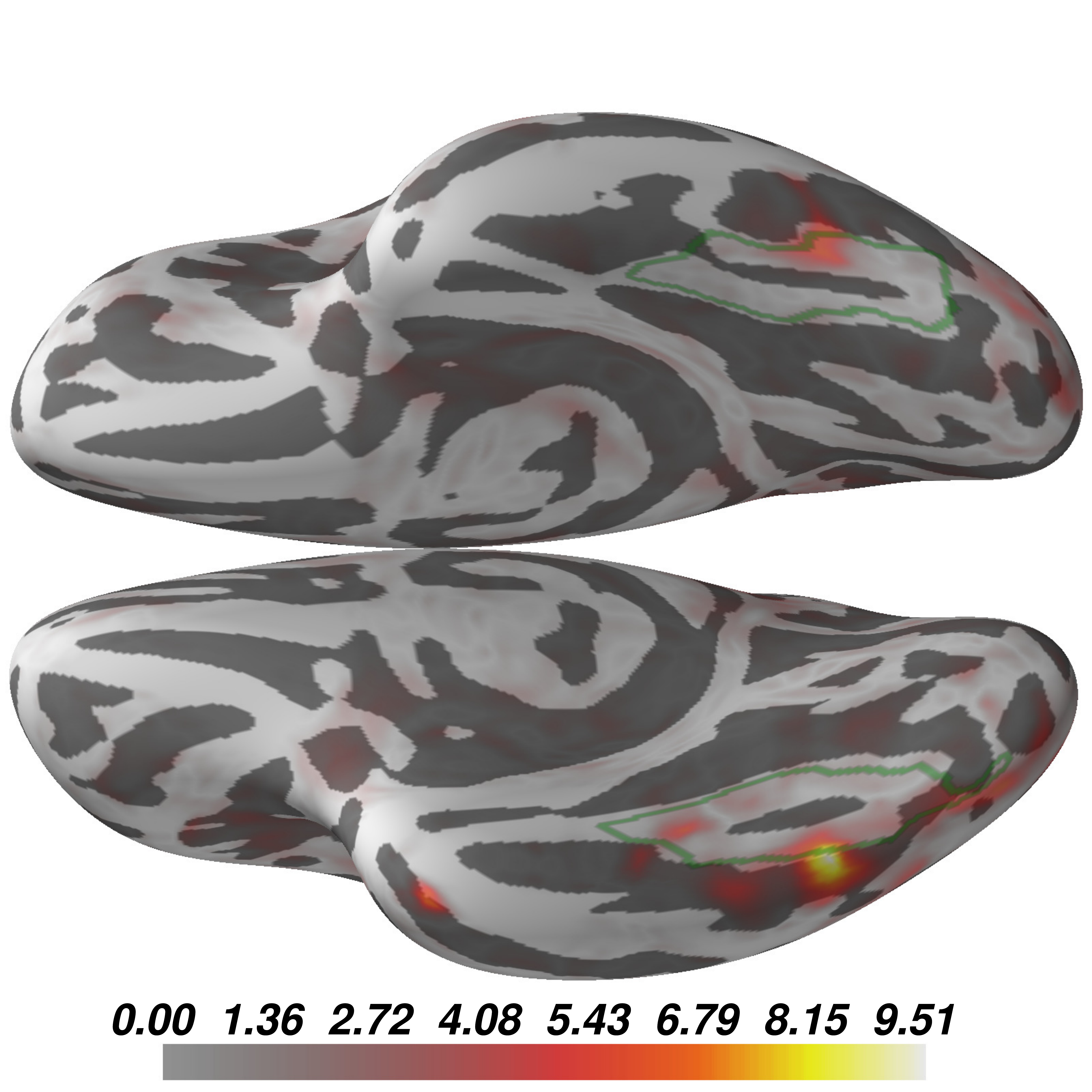}
	\end{minipage}
	\caption{Neural patterns of subject 2. Absolute amplitudes of MEG source estimates (in nAm) given by
		Lasso  (\textbf{Left}) and MWE (\textbf{Middle}).  Absolute values of fMRI Z-scores. (\textbf{Right}). 
	   The fusiform gyrus is highlighted in green. 
		\label{f:subject3}}
\end{figure}
%
The Lasso solutions in Figure~\ref{f:allsubjects} show significant differences between subjects. Since no ground truth exists, one could argue that MWE promotes consistency at the expense of individual signatures. To address this concern we compute the standardized fMRI Z-score of the conditions \emph{famous vs scrambled faces}. We compare Lasso, MWE and fMRI by computing for each subject the ratio
\emph{largest value in fusiform gyrus} / \emph{largest absolute value}. We report the mean across all subjects in Figure~\ref{f:ffg}.
Note that for all subjects, the fMRI Z-score reaches its maximum in the fusiform gyrus, and that MWE regularization leads to more agreement between MEG and fMRI. Figure~\ref{f:subject3} shows MEG with MWE and fMRI results for subject~2.

%% file: sec/conclusion.tex
\section*{Conclusion}
We proposed in this work a novel approach to promote functional consistency through a convex model defined using an Unbalanced Optimal Transport regularization. Using a public MEG and fMRI dataset, we presented experiments demonstrating that MWE outperform multi-task sparse models in both amplitude and support estimation. We have shown in these experiments that MWE can close the gap between MEG and fMRI source imaging by gathering data from different subjects.